\documentclass{article}

\usepackage[final]{nips_2017}

\usepackage[utf8]{inputenc} % allow utf-8 input
\usepackage[T1]{fontenc}    % use 8-bit T1 fonts
\usepackage{hyperref}       % hyperlinks
\usepackage{url}            % simple URL typesetting
\usepackage{booktabs}       % professional-quality tables
\usepackage{amsfonts}       % blackboard math symbols
\usepackage{nicefrac}       % compact symbols for 1/2, etc.
\usepackage{microtype}      % microtypography
\usepackage{graphicx}
\usepackage{subfig}
\newcommand*\samethanks[1][\value{footnote}]{\footnotemark[#1]}
\title{SIMILARnet: Simultaneous Intelligent Localization and Recognition Network}
\author{
Arna Ghosh \thanks{Equal contribution}\\
McGill University\\
Montreal, QC H3A 0G4. Canada.\\
Email: arna.ghosh@mail.mcgill.ca
\And
Biswarup Bhattacharya \samethanks\\
University of Southern California\\
Los Angeles, CA 90089. USA.\\
Email: bbhattac@usc.edu
\And
Somnath Basu Roy Chowdhury \samethanks\\
Indian Institute of Technology\\
Kharagpur, WB 721302. India.\\
Email: brcsomnath@ee.iitkgp.ernet.in
}

\begin{document}
\maketitle
\begin{abstract}
Global Average Pooling (GAP) \cite{GAP} has been used previously to generate class activation for image classification tasks. The motivation behind SIMILARnet comes from the fact that the convolutional filters possess position information of the essential features and hence, combination of the feature maps could help us locate the class instances in an image. We propose a biologically inspired model that is free of differential connections and doesn't require separate training thereby reducing computation overhead. Our novel architecture generates promising results and unlike existing methods, the model is not sensitive to the input image size, thus promising wider application. Codes for the experiment and illustrations can be found at:  {\url{https://github.com/brcsomnath/Advanced-GAP}}.
\end{abstract}

\section{Introduction}
The current advances in computer vision and image recognition have seen wide use of convolutional neural networks (CNNs). In \cite{GAP}, it has been shown that the convolutional units of various layers of CNNs are capable of detecting the position of object in the image without any supervision on the location of object. They show that although earlier convolutional layers are capable of capturing only the low-level features in the image, higher layers are able to capture task-specific features. However, the information about the location of these features in the image is lost when fully connected layers are used for classification. Some fully convolutional networks have been proposed recently such as GoogleNet \cite{googlenet}, and Network in Network \cite{lin2013network} that aim to reduce the number of network parameters while maintaining performance by avoiding the fully-connected layers. 

The work of \cite{lin2013network} introduced the concept of \textit{Global Average Pooling} layers which act as a structural regularizer and prevent overfitting. However, \cite{GAP} shows that the average pooling layers can be used to retain the localization ability of the final layers of the network. This concept is also used in \cite{foodGAP} for localization and binary classification of food items. The main disadvantage in using the Global Average Pooling layers lies in the fact that these models require differential connections during training and deployment.

Looking into the biology of the human nervous system, we know that this is not the case for connections in the brain. Most connections are pruned and stabilized during the development period of the organism. Therefore, we looked into methods to bridge this gap and came up with an architecture where we do not need to alter the connections of a network. The major motivation of our work comes from the fact that different high level features correspond to different classes and the classification of an image to a particular class depends on certain features being excited more heavily than others. Blocking certain specific features affect the perception of humans to perceive or classify the image to a particular class. Drawing from this idea, we believe the errors in classification with a filter absent would be a useful cue to understand the importance of features captured by that filter.

Keeping these in mind, we present a novel architecture that can be used to localize the positions of instances of all classes in an image and then classify each one of those hotspots. Unlike previous methods, the architecture is not as sensitive to the size of the input image and hence has a wider scope of application. The experiments are done on the MNIST dataset and further discussions are presented from the point of view of text extraction from natural images, as a proof of concept, but can be easily extended to other datasets and domains. 

\section{Architecture}
Our work is described using the Le-Net architecture \cite{googlenet}, although the method proposed is easily scalable to any CNN. The network is trained on the MNIST dataset to classify images into $10$ classes (digits in this case). Following the training of the network to obtain a reasonably good classification accuracy, we try to figure out the \textit{importance} of each filter in the last layer for classifying the digits. 

\begin{figure*}[h!]
	\centering
	\includegraphics[width=\textwidth]{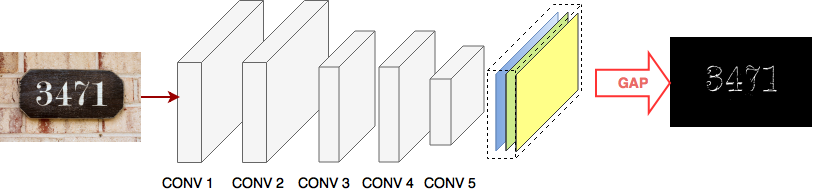}
	\caption{SIMILARnet Architecture}
\end{figure*}

Since the final layer filters have the information about the spatial location of specific features as well, a linear combination of those filters could provide a heatmap of features in the image that are required to be classified into one of the $10$ classes. The idea that a linear combination of the filters would provide an activation map comes from the fact that every super-pixel corresponding to a class instance activates a set of filters responsible for classification of that class. Therefore, a linear combination of the filters would enhance the features that are captured by more filters. However, giving equal weights to all filters leads to information about the filter importance getting dropped. A filter that is responsible for capturing more important features should be given more weight in the linear combination. In \cite{GAP}, these weights are learned by training a separate neural network. However, we try to identify the importance of each filter from the already trained network itself. Since the decision boundary approximated by the network is a piece-wise linear function, it is difficult to understand the importance of the filter from the weights learnt by the network. 

Therefore, we use the error in classification in absence of each filter as a metric of the \textit{importance} of that filter, unlike training another network as in \cite{GAP} and \cite{foodGAP}. We record the error of classification in presence of all the filters as the baseline. The weights for the generation of the heatmap using a linear combination of these filter responses is the difference of error from the baseline. Therefore, if \textbf{w} represents the weight vector for heatmap generation, the overall classification error (error with all filters present) is $E$, and \textbf{e} is the error vector, where $e_i$ represents the error in classification when filter $i$ is missing (or response from filter $i$ is blocked from propagating into further layers), then 
\begin{equation}
w_i' = E - e_i
\end{equation}
The final weights are normalized using softmax operation as shown in Equation (2),
$$w_i = \frac{\exp(w_i')}{\sum\limits_{j=1}^{n} \exp(w_j)}$$ 
where, $n$ is the total number of filters. Suppose the filter response of the final layer filters is represented as \textbf{F} where $\textbf{F}_i$ represents the filter response of the $i^{th}$ filter. Then the heatmap corresponding to the position of the class instances in an image, represented by \textbf{H}, is given by a weighted average of the filter responses. 
\begin{equation}
\textbf{H} = \sum_i w_i \times \textbf{F}_i
\end{equation}
% An illustration of the entire architecture is provided in the \textbf{Figure}. 
Table \ref{net} depicts the parameters of the Le-Net network used for training.

\begin{figure}[h!]
	\centering
	\subfloat[]{\includegraphics[width=2in, height=1.4in]{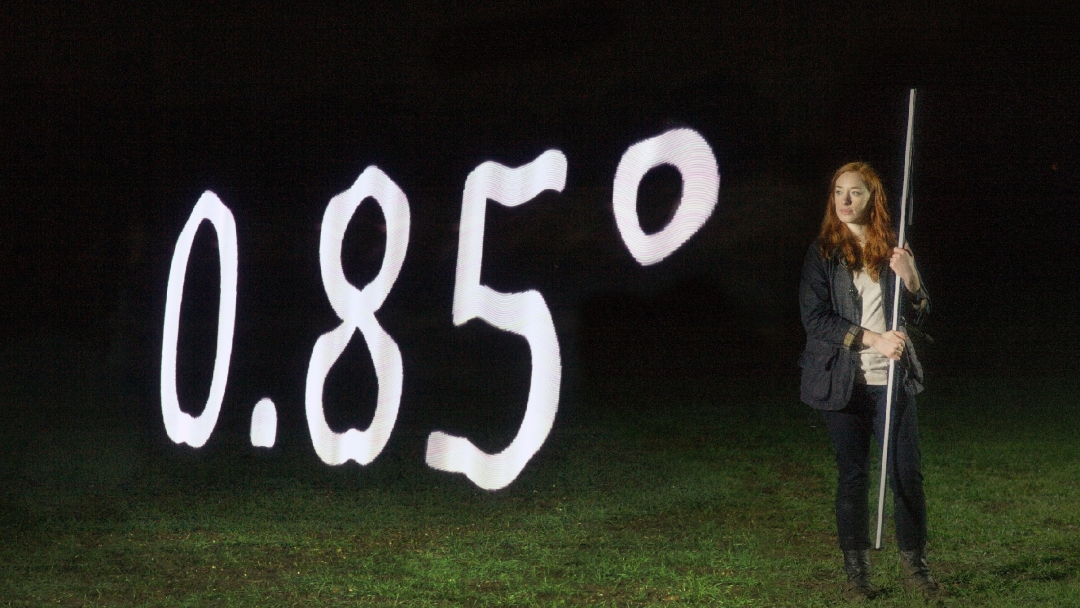}}\hspace*{.05cm}
	\subfloat[]{\includegraphics[width=2in, height=1.4in]{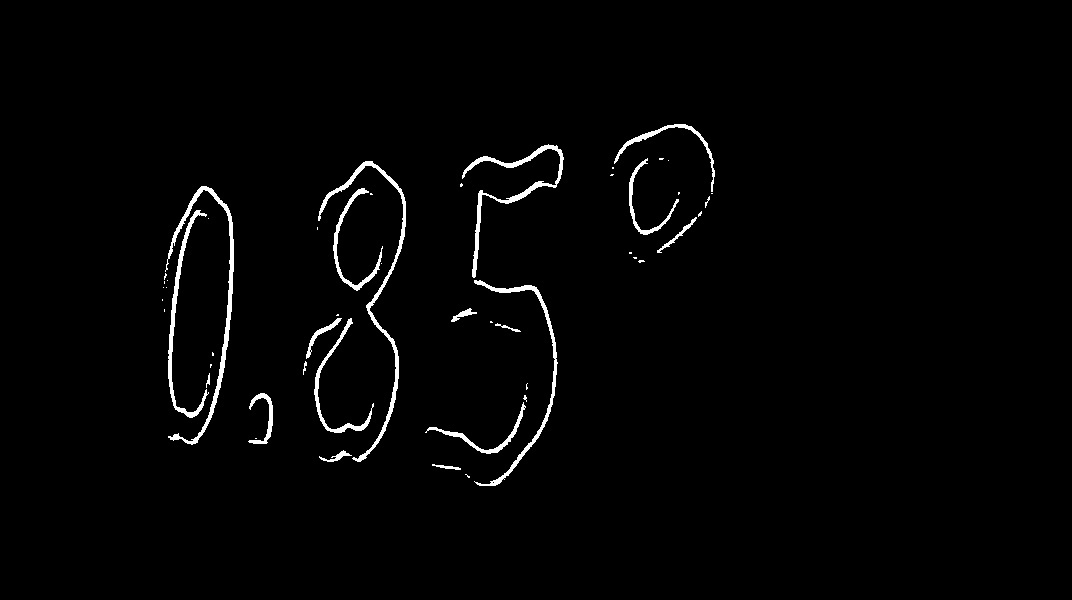}}\\
	\subfloat[]{\includegraphics[width=2in, height=1.4in]{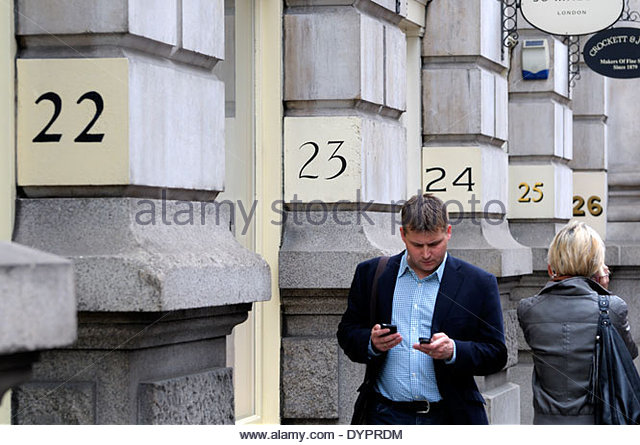}}\hspace*{.05cm}
	\subfloat[]{\includegraphics[width=2in, height=1.4in]{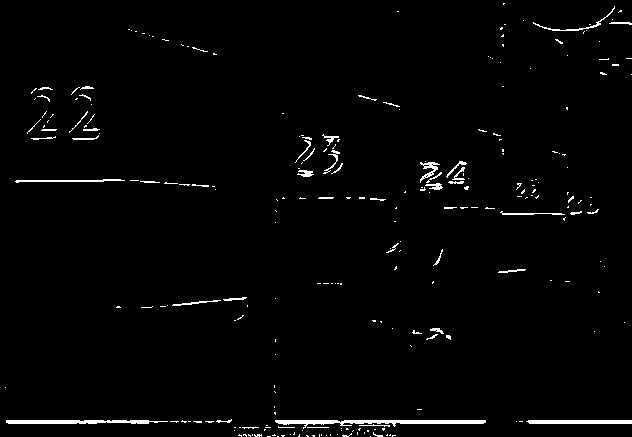}}\\
	\subfloat[]{\includegraphics[width=2in, height=1.4in]{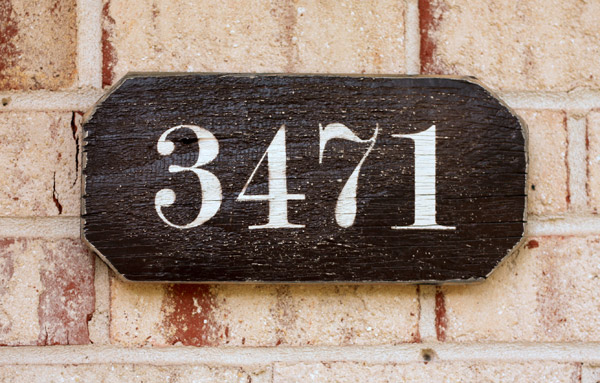}}\hspace*{.05cm}
	\subfloat[]{\includegraphics[width=2in, height=1.4in]{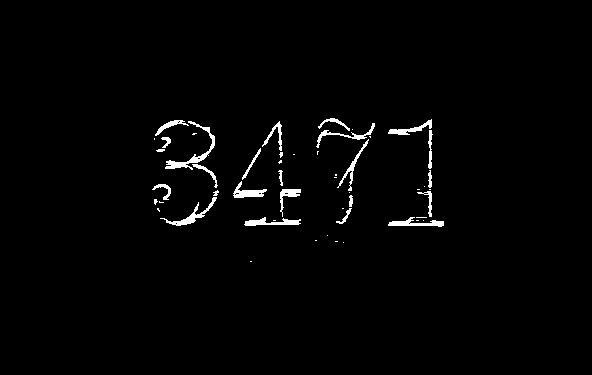}}\\
	\subfloat[]{\includegraphics[width=2in, height=1.4in]{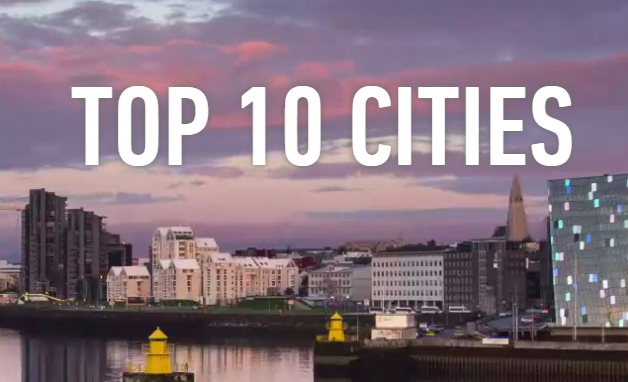}}\hspace*{.05cm}
	\subfloat[]{\includegraphics[width=2in, height=1.4in]{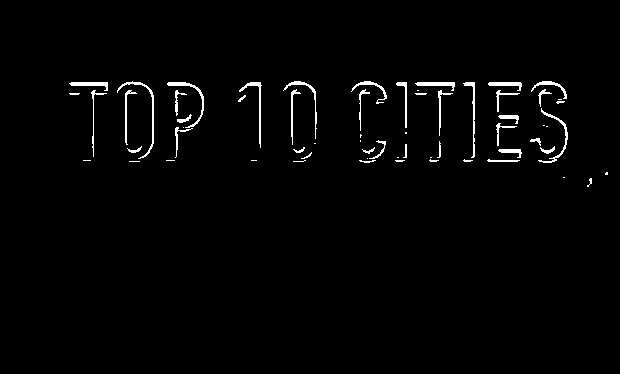}}\\
	\caption{Results of our network (right) on stock photos of different sizes (left)}
	\label{fig:illustration}
\end{figure}

\begin{table}
\centering
\begin{tabular}{l c c c c c}
\hline
Layer & Type & Maps and Neurons & Filter Size\\
\hline
0 & Input & 1M $\times$ 28 $\times$ 28N & - \\
1 & Conv & 6M $\times$ 28 $\times$ 28N & 5$\times$5\\
2 & MaxPool & 6M $\times$ 14 $\times$ 14N & 2$\times$2\\
3 & Conv & 16M $\times$ 14 $\times$ 14N & 5$\times$5\\
4 & MaxPool & 16M $\times$ 7 $\times$ 7N & 2$\times$2\\
5 & FullyConn & 120N & 1$\times$1\\
6 & FullyConn & 84N & 1$\times$1\\
7 & FullyConn & 10N & 1$\times$1\\
\hline
\end{tabular}
\vspace*{0.2cm}
\caption{Network architecture used for digit classification}
\label{net}
\end{table}

\section{Experiments}
\subsection*{Setup}
We use the MNIST handwritten digits dataset for experimentation. We train the network on the MNIST images. The training set has $60000$ images and the test set has $10000$ images. The network is then used to identify the position of class instances (here digits) on natural images. The network is trained using Adam Optimizer with a learning rate of $0.001$ and learning rate decay of $0.00001$. The batch size is kept at $300$ images and the network is trained for $10$ iterations ($200$ batches per epoch). The results show that the network is able to segment the text from the images. 

\subsection*{Results \& Illustrations}
The results are shown in Figure \ref{fig:illustration}. The network is able to extract text and numbers from the images, due to their similarity in features. We use a high threshold value on the heatmap generated from the linear combination of filters to show the extraction property of the network. Although the network is trained on numbers, the network is capable of extracting text as well. This is because the features required to classify numbers and characters are very similar. The class-wise accuracy is mentioned in Table 2.
\begin{table}[h!]
	\centering
	\label{tab:freq}
	\begin{tabular}{ccccl}
		\toprule
		Class & Accuracy & Class & Accuracy\\
		\midrule
		0 & 99.59\% & 5 & 98.32\% \\
		1 & 99.47\% & 6 & 98.23\% \\
		2 & 99.90\% & 7 & 98.64\% \\
		3 & 98.51\% & 8 & 97.43\% \\
		4 & 99.08\% & 9 & 97.92\% \\
		
		\midrule
		& Overall & 98.73\% & \\
		\bottomrule
	\end{tabular}
	\vspace*{0.2cm}
	\caption{Class-wise accuracy}
\end{table}

%\begin{figure}[h!]
%	\centering
%	\includegraphics[scale=0.3]{figures/Training2.png}
%	\caption[Training curve]{Training curve showing the training loss at each batch of forward pass}
%	\label{fig:trainingCurve}
%\end{figure}

\section*{Conclusion}
The proposed network architecture can clearly be used for localization and recognition of trained class instances in an image. The results are presented on different size images to illustrate the versatility of the network. Further experiments would entail deployment on more complex datasets, using deeper networks, and possibly checking for class-specific or category-specific heatmaps to locate instances of objects belonging to similar ``categories'', i.e. objects that respond to similar features.

\bibliographystyle{abbrv}
\bibliography{references}

\begin{thebibliography}{1}

\bibitem{foodGAP}
M.~Bolanos and P.~Radeva.
\newblock Simultaneous food localization and recognition.
\newblock In {\em Pattern Recognition (ICPR), 2016 23rd International
  Conference on}, pages 3140--3145. IEEE, 2016.

\bibitem{lin2013network}
M.~Lin, Q.~Chen, and S.~Yan.
\newblock Network in network.
\newblock {\em arXiv preprint arXiv:1312.4400}, 2013.

\bibitem{googlenet}
C.~Szegedy, W.~Liu, Y.~Jia, P.~Sermanet, S.~Reed, D.~Anguelov, D.~Erhan,
  V.~Vanhoucke, and A.~Rabinovich.
\newblock Going deeper with convolutions.
\newblock In {\em Proceedings of the IEEE conference on computer vision and
  pattern recognition}, pages 1--9, 2015.

\bibitem{GAP}
B.~Zhou, A.~Khosla, A.~Lapedriza, A.~Oliva, and A.~Torralba.
\newblock Learning deep features for discriminative localization.
\newblock In {\em Computer Vision and Pattern Recognition (CVPR), 2016 IEEE
  Conference on}, pages 2921--2929. IEEE, 2016.

\end{thebibliography}

\end{document}